\documentclass[sigconf]{aamas} 

\usepackage{balance} 
\usepackage{rotating}
\usepackage{graphicx}
\usepackage{lscape}
\usepackage{amsmath}
\usepackage{float}
\usepackage{multirow}
\usepackage{algorithm}
\usepackage{algpseudocode}
\usepackage{tcolorbox}

\setcopyright{ifaamas}
\acmConference[AAMAS '24]{Proc.\@ of the 23rd International Conference
on Autonomous Agents and Multiagent Systems (AAMAS 2024)}{May 6 -- 10, 2024}
{Auckland, New Zealand}{N.~Alechina, V.~Dignum, M.~Dastani, J.S.~Sichman (eds.)}
\copyrightyear{2024}
\acmYear{2024}
\acmDOI{}
\acmPrice{}
\acmISBN{}

\acmSubmissionID{390}

\title[Robust Knowledge Extraction from Large Language Models using Social Choice Theory]{Robust Knowledge Extraction from Large Language Models using Social Choice Theory}

\author{Nico  Potyka}
\affiliation{
  \institution{Cardiff University, UK}
  \city{ }
  \country{ }}
\email{potykan@cardiff.ac.uk}
\authornotemark[1]

\author{Yuqicheng Zhu}
\affiliation{
  \institution{Bosch Center for AI, Germany}
  \city{Univ. of Stuttgart, Germany}
  \country{}}
\email{Yuqicheng.Zhu@de.bosch.com}
\authornote{Equal Contribution.}

\author{Yunjie He}
\affiliation{
  \institution{Bosch Center for AI, Germany}
  \city{Univ. of Stuttgart, Germany}
  \country{}}
\email{Yunjie.He2@de.bosch.com}

\author{Evgeny Kharlamov}
\affiliation{
  \institution{Bosch Center for AI, Germany}
  \city{Univ. of Oslo, Norway}
  \country{}
  }
\email{Evgeny.Kharlamov@de.bosch.com}

\author{Steffen Staab}
\affiliation{
  \institution{Univ. of Stuttgart, Germany}
  \city{Univ. of Southampton, UK}
  \country{}}
\email{steffen.staab@ki.uni-stuttgart.de}

\begin{abstract}
Large-language models (LLMs) can support
a wide range of applications like conversational agents,
creative writing or general query answering. However, they are ill-suited for 
query answering in high-stake domains like medicine
because they are typically not
robust - even the same query can result in different answers when prompted
multiple times. In order to improve the robustness of LLM queries, 
we propose using ranking queries repeatedly and to aggregate
the queries using methods from social choice theory. 
We study ranking queries in diagnostic settings like medical and
fault diagnosis and discuss how the
Partial Borda Choice function from the literature can be applied
to merge multiple query results. We discuss some additional interesting
properties in our setting and evaluate the robustness of our 
approach empirically.
\end{abstract}


\newcommand{\BibTeX}{\rm B\kern-.05em{\sc i\kern-.025em b}\kern-.08em\TeX}
\newcommand{\down}{\ensuremath{\mathrm{Down}_\succeq}}
\newcommand{\downi}{\ensuremath{\mathrm{Down}_{\succeq_i}}}
\newcommand{\inc}{\ensuremath{\mathrm{Inc}_\succeq}}

\newcommand{\fpbw}{\ensuremath{f^\textit{PBW}}}
\newcommand{\spbw}{\ensuremath{s^\textit{PBW}}}
\newcommand{\spbwn}{\ensuremath{\overline{s}^\textit{PBW}}}
\newcommand{\pbw}{\ensuremath{w^\textit{PBW}}}

\makeatletter
\gdef\@copyrightpermission{
	\begin{minipage}{0.3\columnwidth}
		\href{https://creativecommons.org/licenses/by/4.0/}{\includegraphics[width=0.90\textwidth]{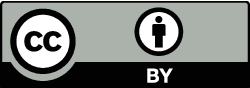}}
	\end{minipage}\hfill
	\begin{minipage}{0.7\columnwidth}
		\href{https://creativecommons.org/licenses/by/4.0/}{This work is licensed under a Creative Commons Attribution International 4.0 License.}
	\end{minipage}
	\vspace{5pt}
}
\makeatother

\begin{document}


\pagestyle{fancy}
\fancyhead{}


\maketitle 


\section{Introduction}

Large Language Models (LLMs) achieve state-of-the-art results in various natural language processing (NLP) tasks. Formally, LLMs represent a conditional probability distribution $P(T_{n+1} | T_1, \dots, T_n)$  over tokens (character sequences) that predicts the next token given a fixed context of previous tokens. 
To answer a query $Q$, $Q$ is decomposed into tokens $T_1, \dots, T_Q$ and used to sample the first token $A_1$ of the answer from $P(A_1 | T_1, \dots, T_Q)$. $A_1$ can then be added to the context $T_1, \dots, T_Q$ to sample the next answer token $A_2$. This process is repeated until a special end of text token is reached. Since finding an optimal sequence of answer tokens is hard,
answer sequences are  often
computed by a heuristic search like Beam \cite{huang2001spoken}, Top-K \cite{Fan18} or Nucleus \cite{HoltzmanBDFC20} search that build up multiple promising token sequences in
parallel.
The sampling process is controlled by a temperature parameter. For temperature $0$, the algorithms samples greedily. Increasing the temperature allows sampling tokens with lower
local probability. While some authors associate higher
temperatures with more creative answers, they can also result in higher probability answers because greedy selection can exclude high probability
sequences that start with low probability tokens. 

Given the success of LLMs in difficult NLP taks, they are increasingly being used for general question answering tasks. This is a natural application as it is reasonable to assume that LLMs picked up a lot of interesting information during training. However, one limitation of LLMs is that they will always produce an answer even if they did not learn anything
about the question. This problem is referred to as
\emph{hallucination} in the literature \cite{ji2023survey}.
The uncertainty of an answer is hard to quantify.
While every answer sequence can be associated with a probability, this is merely the probability of the text sequence and should not be confused with the probability that the answer is correct
(or that the LLM "believes" that the answer is correct). Theoretically, LLMs can be asked to output probabilities for their predictions, but
it is hard to say how meaningful these probabilities are since there 
is nothing in a typical LLM architecture that would allow them to 
infer meaningful probabilities (unless they picked up a particular probability from the training corpora). Since query answering with
LLMs is based on a heuristic search for high probability token sequences
rather than on reasoning, we, in particular, have the following types
of uncertainty:
\begin{enumerate}
    \item Query-Uncertainty: prompting the same query repeatedly can result in different answers.
    \item Syntax-Uncertainty: semantically equivalent queries that differ only syntactically can result in different answers.
    \item Distraction-Uncertainty: meaningless information added to
    the query can result in a different answer.
\end{enumerate}
Let us note that, in principle, query uncertainty can be eliminated by
setting the temperature parameter to $0$. However, as outlined above,
the deterministic answer will be somewhat random because it corresponds
to some local optimum found by a heuristic search algorithm.
We therefore aim at allowing some randomness in the answer, but
increasing the robustness. The idea of robustness is that similar
queries should result in similar answers. In particular, the same
query prompted multiple times can result in different answers in
our setting. However, we would like that these answers are semantically
similar. Similarly, we would like that syntactic changes of a query do not
change the semantics of the answer.

In this work, we will mainly focus on making LLMs robust against
\emph{query uncertainty}, but we will also look at \emph{syntax uncertainty} in our experiments. We explore to which extent an answer sampling strategy combined with social choice theory methods can improve the robustness of LLMs. The idea is as follows: instead of asking a query once, we ask it repeatedly (starting each time from the original question context). Our assumption is that if the LLM picked up the answer during training, then this answer should occur in the majority of cases. On the other hand, if it did not pick up information about the query and hallucinates an answer, we expect that the different answers will be very random. We will apply tools from social choice theory to aggregate the answers. We expect that, if the
LLM picked up meaningful information, then our aggregation will result
in a clear ranking of the different answers,
while 
it will be mostly indifferent
between the answers otherwise.

Let us emphasize that the outcome
should be interpreted with care. If the LLM has been trained on a
text corpora with false information, we may find that an LLM gives
a false answer with high certainty. The probabilities that we derive
should therefore be understood as subjective probabilities that 
reflect the uncertainty of the LLM and not as statistical probabilities. 
We view our method as most useful when being applied to LLMs that 
were trained on reliable literature 
(e.g., peer-reviewed articles and books) and not 
on random text from the internet.
While reliable pretrained models like BioBert and MedBert exist \cite{lee2020biobert,rasmy2021med}, they still require fine-tuning
to be usable as question answering systems.
Since our resources are limited,
we will therefore use ChatGPT-turbo in our experiments, which was trained on mixed data with varying
reliability. However, our experiments are only a proof of concept
and the idea can directly be transferred to LLMs trained on
high quality domain-specific data.

In our investigation, we will focus on diagnostic problems,
where we try to identify the cause of a particular situation
or condition. The identified cause is called the \emph{diagnosis
} for the condition. Typical examples are medical diagnosis (identify the medical condition that causes a set of symptoms) or fault diagnosis (identify the defective component in a technical system that causes malfunctions). The query consists of a description of the situation and we ask
for a ranking of possible causes ordered by their plausibility.
In order to take account of uncertainty, we repeat the query multiple times
and collect the rankings. Tools from social choice theory can then be
applied to merge the rankings and to quantify the uncertainty of the 
answer. To do so, we will build up on scoring-based voting methods
for partial preference orderings \cite{cullinan2014borda}.

\section{Related Work}

Prior research on uncertainty quantification of LLMs focused on investigating the 
probabilities of token sequences
\cite{guo2017calibration,jiang2021can}. However,
as discussed before, the probability of the token sequence
should not be confused with the probability that the token 
sequence expresses a valid claim.
In particular, the same claim can be expressed by different (semantically equivalent) token sequences that obtain different probabilities.
\cite{kuhn2023semantic} address this issue by first clustering claims with the same semantic meaning and summing their probabilities to calculate a "semantic entropy". Other work involves training or fine-tuning the LLMs to quantify uncertainty \cite{kadavath2022language, lin2022teaching, mielke2020linguistic}. However, due to lack of transparent training specifics, these approaches might be difficult to reproduce in addition to being expensive.

Despite the demand for uncertainty quantification without relying on model fine-tuning or accessing the proprietary information of LLMs, there is little work in this area and much remains unexplored. To our best knowledge, only \cite{xiong2023can, tian2023just} quantify uncertainty based on the verbalized confidence given by LLMs or self-consistency of the claims. The significance of verbalized confidence is unclear since there is nothing in a typical LLM architecture that would allow it to infer meaningful probabilities. Our approach aggregates answers 
and quantifies the uncertainty using methods from social choice theory. Moreover, We study queries that give a rank with multiple possible answers rather than one single answer as an output, no approach from existing work can be directly applied in our case.

The recent neuro-symbolic theorem prover \emph{LINC} \cite{OlaussonGLZSTL23} uses LLMs as a semantical parser
to translate natural language reasoning problems into
first-order logic that can then be processed by a 
symbolic theorem prover. To decrease the risk of
parsing errors, the authors parse and process the inputs
repeatedly and apply majority voting to determine the outcome.
This may be another interesting domain for applications
of more sophisticated voting methods.

The notion of robustness that we consider here (similar inputs
should result in similar outputs) follows the terminology in
\emph{Explainable AI} \cite{Alvarez2018,JMLR:v24:23-0042,leofante24}
and should not be confused with statistical \cite{Huber81}
or adversarial \cite{SzegedyZSBEGF13} robustness. From
an explanation point of view,
our scoring method is interpretable in the sense that
the scores can be explained from the LLM's responses to the repeated prompts. The responses can be further explained
by the LLM's sampling procedure and the output probabilities of the transformer. However, understanding the output probabilities
of transformers is difficult and a topic of current research
\cite{zhao2023explainability}.

\section{Social Choice Theory Background}

Social choice theory deals with aggregating
individual preferences of different agents
towards a collective choice 
\cite{socialchoice2016}.
The agents are often seen as voters who
can express their preferences in different
ways. For example, they may be able to 
vote for a single candidate, for multiple
candidates or report a preference ordering
over the candidates. We will focus on the
latter setting here. 
Formally, we consider a finite set of \emph{voters}
$N = \{1, \dots, n\}$ and a finite
set of outcomes $O = \{o_1, \dots, o_m\}$.
A \emph{partial order} $\succeq$ over $O$ is a binary 
relation over $O$ that is reflexive,
anti-symmetric and transitive.  We do not
assume that it is complete, that is, there can
be outcomes $o_i \neq o_j$ such that
neither 
$o_i \succeq o_j$ nor $o_j \succeq o_i$.
As usual, we write 
\begin{itemize}
    \item $o \succ o'$ iff $o \succeq o'$ and
$o' \not\succeq o$, 
    \item $o \sim o'$ iff 
$o \succeq o'$ and
$o' \succeq o$.
\end{itemize}
If $o \succ o'$, we say that $o$ is \emph{strictly preferred} to $o'$
and if $o \sim o'$, we say that we are \emph{indifferent} between the two.
A \emph{profile} $p = [\succeq_1, \dots, \succeq_n]$ contains one partial order
for every voter and captures
the preferences expressed by them.

The process of aggregating the voters' preferences 
can be formalized in different ways.
A \emph{social choice function} is a mapping $f$ from the
set of all profiles to a non-empty subset of the outcomes. 
Intuitively, $f(p)$ should contain the outcomes that are
maximally preferred by the voters. Ideally, $f(p)$ contains
only a single element, but there are cases where a unique choice
cannot be made without any ad-hoc assumptions (like chosing 
a random outcome or a lexicographically minimal one).


Social choice research
often focuses on total orderings, where agents
express preferences over all possible
outcomes \cite{socialchoice2016}. In our application, the 
outcomes are possible diagnoses, 
and the different answers do not necessarily
contain the same diagnoses. We will
therefore focus on preferences expressed by partial orderings.
Since we are interested in quantifying the uncertainty of
an answer (based on the variance in the rankings),
scoring-based voting methods are a natural choice.
We recall some ideas about aggregating partial preferences
by scoring-based voting methods from \cite{cullinan2014borda}.

To begin with, a \emph{scoring procedure} $s_p: O \rightarrow \mathbb{R}$ is a mapping from outcomes to numerical values 
that is parametrized by a profile $p$ \cite{cullinan2014borda}.
Intuitively, $s_p(o)$ is the score of outcome $o$
with respect to the preferences expressed by the profile $p$.
Every scoring procedure induces a social choice function by letting
\begin{equation}
\label{def_f_from_s}
    f(p) = \arg \max_{o \in O} s_p(o). 
\end{equation}
A \emph{weighting procedure} $w_\succeq: O \rightarrow \mathbb{R}$ maps outcomes to numerical values and is
parametrized by a partial order $\succeq$ \cite{cullinan2014borda}.
Intuitively, $w_\succeq(o)$ is the score of outcome $o$
with respect to the preferences expressed by $\succeq$.
We can construct a scoring procedure from a weighting procedure
by letting \cite{cullinan2014borda}
\begin{equation}
\label{def_s_from_w}
s_p(o) = \sum_{i=1}^n  w_{\succeq_i}(o),
\end{equation}
where we assume  $p = [\succeq_1, \dots, \succeq_n]$.

A weighting procedure, in turn, can be based on how many other
outcomes are less preferred and how many are incomparable.
To do so, we can consider functions $\down: O \rightarrow 2^O$
and $\inc: O \rightarrow 2^O$ defined as follows
\cite{cullinan2014borda}:
\begin{align}
& \down(o) = |\{ o' \in O \mid o \succ o' \}|, \\
& \inc(o) = |\{ o' \in O \mid \textit{$o$ and $o'$ are incomparable} \}|,   
\end{align}
where, for a set $S$, $|S|$ denotes its cardinality.
That is, $\down(o)$ is the number of outcomes ranked lower than
$o$ and $\inc(o)$ is number of outcomes incomparable to $o$.

The following two properties of weighting procedures have been
proposed in \cite{cullinan2014borda}:
\begin{description}
    \item[Linearity:] There exist constants $\alpha, \beta, \gamma \in \mathbb{R}$ such that 
    \begin{equation}
    \label{eq_linearity}
     w_\succeq(o) = \alpha \cdot \down(o) + \beta \cdot \inc(o) +
    \gamma.   
    \end{equation}
    \item[Constant Total Weight:] There exists a constant $\delta$ such
    that $\sum_{o \in O} w_\succeq(o) = \delta$ for all partial orders.
\end{description}

 \emph{Partial Borda Weighting (PBW)} \cite{cullinan2014borda}
 $\pbw_\succeq$
 is the linear weighting procedure defined by letting
 \begin{equation}
     \alpha=2, \beta = 1, \gamma = 0
 \end{equation}
in \eqref{eq_linearity}. 
\begin{definition}[PBW Weighting] The \emph{PBW weighting} procedure
is defined as 
    \begin{equation}
         \pbw_\succeq(o) = 2 \cdot \down(o) + \inc(o) .   
    \end{equation}
\end{definition}
One can show the following.
\begin{theorem}[\cite{cullinan2014borda}]
\begin{enumerate}
$\pbw_\succeq$ satisfies Linearity
and Constant Total Weight and every other weighting procedure
that satisfies these two properties is an affine transformation
of PBW.   
\end{enumerate}
\end{theorem}
We refer to \cite[Theorem 1]{cullinan2014borda} for more
details about this result.

The \emph{partial Borda choice function} $\fpbw$ is the 
social choice function induced by $\pbw_\succeq$ based on
equations \eqref{def_f_from_s} and 
\eqref{def_s_from_w}. It can be characterized
as follows.
\begin{theorem}[\cite{cullinan2014borda}]
\label{theorem_pbc_properties}
The partial Borda choice function is the unique social choice 
function that satisfies the following properties.
\begin{description}
    \item[Consistency:] If $p_1, p_2$ are disjoint profiles
    and $f(p_1) \cap f(p_2) \neq \emptyset$ then
    $f(p_1) \cap f(p_2) = f(p_1 \cup p_2)$.
    \item[Faithfulness:] If $p = [\succeq_1]$ and
    $b \succeq_1 a$, then $a \not \in f(p)$.
    \item[Neutrality:] $f$ is invariant with respect to permutations
    of $O$ (renaming the outcomes will not affect the result), that is,
    $f( \sigma(p)) = \sigma( f(p))$ for all bijective mappings 
    $\sigma: O \rightarrow O$.
    \item[Cancellation:] If for all outcomes $o_1 \neq o_2$, the
    number of voters who rank $o_1$ above $o_2$ equals the
    number of voters who rank $o_2$ above $o_1$, then
    $f(p) = O$.
\end{description}
\end{theorem}
We refer to \cite[Theorem 2]{cullinan2014borda} for more
details about this result.




\section{Improving the Robustness of LLM Queries with PBW}

As we saw in the previous section, aggregating partial preferences
with PBW gives us several desirable analytical guarantees.
We will now use PBW to improve the robustness of LLM ranking 
queries. The basic idea is to ask the LLM for the most plausible 
explanations of a situation repeatedly and to use PBW to
aggregate the answers.

\subsection{From Queries to Rankings}

In order to obtain ranking answers from LLMs, we consider queries
of a special form that we call \emph{ranking queries}. We refrain from
a formal definition and just explain the intuitive idea. Roughly speaking,
a ranking query consists of
\begin{itemize}
    \item a \emph{condition description},
    \item \emph{answer instructions}. 
\end{itemize}
\begin{example}
As a running example, we will use a medical scenario with the following ranking query:
\begin{tcolorbox}
    \begin{quote}
 \textit{"A 20 year old professional runner suffers from a stinging pain in the forefoot. The foot is swollen and stiff. What are the most plausible explanations? Please keep the answer short and order by decreasing plausibility."}
\end{quote}
\end{tcolorbox}
The first two sentences describe the condition, the last two sentences give
the answer instructions. A typical answer provided by ChatGPT looks as follows:
\begin{quote}
The most plausible explanations for a 20-year-old professional runner experiencing a stinging pain, swelling, and stiffness in the forefoot, ordered by decreasing plausibility, could be:
\begin{itemize}
    \item Overuse Injury: Repetitive stress from running may have led to an overuse injury such as metatarsalgia or stress fracture.
    \item Tendonitis: Inflammation of tendons in the forefoot, like extensor tendinitis, could cause these symptoms.
    \item Ligament Sprain: A sprained ligament, like in the metatarsophalangeal joint, could result in pain, swelling, and stiffness.
    \item Footwear Issues: Poorly fitting or worn-out running shoes may contribute to foot problems and discomfort.
    \item Infection or Insect Bite: Infection or an insect bite should be considered but is less likely without specific signs like redness and fever. 
\end{itemize}
\end{quote}
\end{example}
In order to obtain our rankings, we run the prompt 5 times, each time 
starting from an empty context. 
The possible causes provided in the 5 answers, define our set of outcomes $O$.
We summarize and normalize the answers such that synonyms and syntactic differences do not lead to different outcomes.
Given an answer list $o_1, \dots, o_q$ for one prompt, we associate
it with the partial ordering
\begin{equation}
o_1 \succ \dots \succ o_q \succ \overline{\{o_1, \dots, o_q\}},    
\end{equation}
where for every subset $S \subseteq O$, $\overline{S} = O \setminus S$
denotes the complement of $S$ and $o \succ S$ is short for
$o \succ o'$ for all $o' \in S$.
That is, the outcomes occuring in the answer are preferred according
to their order of appearance and they all are preferred to those outcomes
that have not occured. The outcomes that did not occur are incomparable
with respect to this ranking.
\begin{example}
For our running example, we obtained the following outcomes after
manual normalization:
\begin{enumerate}
    \item $\mathrm{bu}$: bursitis,
    \item $\mathrm{fi}$: footwear issues,
    \item $\mathrm{go}$: gout,
    \item $\mathrm{in}$: infection,
    \item $\mathrm{mn}$: Morton's neuroma,
    \item $\mathrm{msr}$: metatarsal stress reaction,
    \item $\mathrm{ni}$: neurological issue,
    \item $\mathrm{oi}$: overuse injury,
    \item $\mathrm{pf}$: plantar fasciitis,
    \item $\mathrm{sf}$: stress fracture,
    \item $\mathrm{te}$: tendonitis,
    \item $\mathrm{tr}$: trauma.
\end{enumerate}
The 5 answers for our running example correspond to the following partial oderings:
\begin{align*}
& \mathrm{oi} \succ_1 \mathrm{fi} \succ_1 \mathrm{tr} \succ_1 \mathrm{in} \succ_1 \mathrm{ni} \succ_1 \overline{\{\mathrm{oi}, \mathrm{fi}, \mathrm{tr}, \mathrm{in}, \mathrm{ni}\}}, \\
& \mathrm{oi} \succ_2 \mathrm{pf} \succ_2 \mathrm{fi} \succ_2 \mathrm{in} \succ_2 \mathrm{go} \succ_2  \mathrm{tr} \succ_2 \overline{\{\mathrm{oi}, \mathrm{pf}, \mathrm{fi}, \mathrm{in}, \mathrm{go}, \mathrm{tr}\}}, \\
&  \mathrm{oi} \succ_3 \mathrm{tr} \succ_3 \mathrm{fi} \succ_3 \mathrm{in} \succ_3 \overline{\{\mathrm{oi}, \mathrm{tr}, \mathrm{fi}, \mathrm{in}\}}, \\
&  \mathrm{sf} \succ_4 \mathrm{pf} \succ_4 \mathrm{mn} \succ_4 \mathrm{msr} \succ_4 \mathrm{bu,} \succ_4 \overline{\{\mathrm{sf}, \mathrm{pf}, \mathrm{mn}, \mathrm{msr}, \mathrm{bu}\}} \\
&  \mathrm{oi} \succ_5 \mathrm{te} \succ_5 \mathrm{fi} \succ_5 \mathrm{in} \succ_5 \overline{\{\mathrm{oi}, \mathrm{te}, \mathrm{fi}, \mathrm{in}\}} . \\
\end{align*}
\end{example}
We constructed the partial orders in our running example manually.
In our experiments, we will use a more automated process
that works as follows:
\begin{enumerate}
    \item \emph{Determine Base-Outcomes: } Query the LLM for a list of potential causes 
    that 
    we call \emph{base-outcomes}.
    \item \emph{Determine Rankings: } Repeatedly ask the LLM for the most plausible causes and to rank them by their plausibility. We call these outcomes \emph{ranking-outcomes}.
    \item \emph{Normalize Rankings: } Normalize the rankings
    by matching ranking-outcomes with
    base-outcomes.
    We use word embeddings (Sentence-BERT\cite{reimers2019sentence}) to map the ranking-outcomes to the 
    most similar base-outcomes. If the similarity of a
    ranking-outcome to all base-outcomes is smaller than
    $0.5$, it will be discarded (and reported).
\end{enumerate}
One can think of other methodologies to compute rankings from LLMs.
To abstract from the details, let us assume that we
have a transformation method $T(Q,N, t)$ of the following form.
\begin{definition}
A \emph{transformation method} $T(Q, N, t)$ takes a ranking query as input,
prompts it $N$ times and produces a profile $[\succeq_1, \dots, \succeq_N]$
from the answer rankings. The parameter $t$ represents the time at which
the query has been prompted.
\end{definition}
The time parameter $t$ is only a technical device to
take account of the fact that the output of LLMs is non-deterministic.
It can also be seen as the (unknown) random seed of the LLM.
The time parameter allows us talking about potentially different outputs
when aggregating repeatedly for the same input. For example, 
say we aggregate the answers for $Q$ five times and then again five
times, then we can denote the two results by $T(Q, 5, t_1)$ and
$T(Q, 5, t_2)$. We will use this notation for the discussion of the
\emph{consistency} property later. The notation is also useful to make 
the idea of robustness more precise. Assume that we have an aggregation
method $A$ that aggregates the profiles obtained from a transformation
method $T$ in some way. Roughly speaking, we say that a pair $(T,A)$ consisting
of a transformation method $T$ and an aggregation method $A$ is 
\begin{description}
    \item[query-robust] if the answers obtained for one query $Q$
    from $A(T(Q, N, t_1))$ and $A(T(Q, N, t_2))$ are "similar" when $N$ is chosen sufficiently large,
    \item[syntax-robust] if the answers for two
    syntactically different, but semantically similar queries $Q_1, Q_2$ from $A(T(Q_1, N, t_1))$ and $A(T(Q_2, N, t_2))$ are "similar" when $N$ is chosen sufficiently large.
\end{description}
The choice of the similarity measure depends on the application.
Correlation measures seem to be a natural choice for measuring
similarity between rankings. Measuring similarity between queries is more difficult. For experiments, one simple way
to generate similar queries is to make purely syntactical
changes to a base query to obtain (almost) semantically
equivalent queries. 

In our application, our aggregation method $A_{PBW}$ ranks the diagnoses from the given profile
by their PBW score. We will use correlation measures to determine the similarity of these rankings for $N=5$ in our experiments.
\subsection{Answer Aggregation}

In order to quantify the plausibility of different answers, we apply
the PBW score. The larger the score, the more plausible the answer.
To make the interpretation of the scores easier, we normalize them
such that all values are between $0$ and $1$. We let
\begin{equation}
    \spbwn(o) = \frac{\spbw(o)}{\sum_{o' \in O} \spbw(o')}
\end{equation}
\begin{table}[]
    \centering
    \caption{PBW scores for running example: first column shows, the outcomes, columns 2-6 show the partial PBW scores per ranking, column 7 shows the PBW scores and column 8 the normalized PBW scores rounded to two digits.}
    \label{tab:exp_runner_scores}
    \begin{tabular}{lccccccc}
         $O$ & $\succ_1$ & $\succ_2$ & $\succ_3$ & $\succ_4$ & $\succ_5$
          & $\spbw(o)$ & $\spbwn(o)$\\
         \hline
    $\mathrm{bu}$ & 6  & 5  & 7  & 14 & 7  & 39 & 0.06\\ 
    $\mathrm{fi}$ & 20 & 18 & 18 & 6  & 18 & 80 & 0.12\\
    $\mathrm{go}$ & 6  & 14 & 7  & 6  & 7  & 40 & 0.06\\
    $\mathrm{in}$ & 16 & 16 & 16 & 6  & 16 & 70 & 0.10\\
    $\mathrm{mn}$ & 6  & 5  & 7  & 18 & 7  & 43 & 0.07\\
    $\mathrm{msr}$ & 6 & 5  & 7  & 16 & 7  & 41 & 0.06\\
    $\mathrm{ni}$ & 14 & 5  & 7  & 6  & 7  & 39 & 0.06\\
    $\mathrm{oi}$ & 22 & 22 & 22 & 6  & 22 & 94 & 0.14\\
    $\mathrm{pf}$ & 6  & 20 & 7  & 20 & 7  & 60 & 0.09\\
    $\mathrm{sf}$ & 6  & 5  & 7  & 22 & 7  & 47 & 0.07\\
    $\mathrm{te}$ & 6  & 5  & 7  & 6  & 20 & 44 & 0.07\\
    $\mathrm{tr}$ & 18 & 12 & 20 & 6  & 7  & 63 & 0.10
    \end{tabular}
\end{table}
Table \ref{tab:exp_runner_scores} shows the PBW scores for our running example.

\subsection{Properties}

We now discuss some analytical guarantees of our approach.
Let us note that the normalized PBW score $\spbwn$ is just a rescaling
of the PBW score $\spbw$. Therefore, the outcomes with maximal score 
and their relative order remains unchanged. To begin with, let us 
reinterpret the properties from Theorem \ref{theorem_pbc_properties}
in our setting.
\begin{description}
    \item[Consistency:] Let $Q$ be a ranking query and let $p_1 = T(Q, N_1, t_1)$, $p_2 = T(Q, N_2, t_2)$. If $o$ has maximum score with respect to both
    $\spbwn(p_1)$ and $\spbwn(p_2)$, then $o$ also has maximum score
    with respect to $\spbwn(p_1 \cup p_2)$.
    \item[Faithfulness:] If we prompt the query only once, then the highest
    ranked outcome obtains the maximum score. 
    \item[Neutrality:] The score of outcomes is independent of their identity.
    \item[Cancellation:] If for all outcomes $o_1 \neq o_2$, the
    number of rankings that rank $o_1$ above $o_2$ equals the
    number of rankings that rank $o_2$ above $o_1$, then
    all outcomes get the same score.
\end{description}
As explained before, the above properties are sufficient to characterize PBW scoring \cite{cullinan2014borda}. That is, there is no other scoring function
that satisfies all these properties (up to affine transformations).
Since all properties seem desirable in our setting, $\spbwn$ is a natural
choice. In the following proposition, we note some additional desirable properties of $\spbwn$ and $\spbw$ in our setting.
The properties also hold for other instantiations of \eqref{eq_linearity}
as long as $\alpha > \beta$ remains satisfied. 
\begin{proposition} Let $Q$ be a query that was prompted $N$ times and resulted 
in the outcomes $O$ and profile $p = [\succ_1, \dots, \succ_N]$.
\begin{description}
    \item[Partial Agreement:] If there are $o_1, o_2 \in O$ such that
    $o_1 \succ_i o_2$ for all $1 \leq i \leq N$, then 
    $\spbwn(o_1) > \spbwn(o_2)$.
    \item[Full Agreement:] If prompting the query repeatedly resulted in
    the same rankings, that is, $\succ_i \ = \ \succ_j$ for all $1 \leq i < j \leq N$
    , then 
    $\spbwn(o_1) > \spbwn(o_2)$ if and only if $o_1 \succeq_i o_2$. 
    \item[Domination:] If there is an $o^* \in O$ such that
    $o^* \succ_i o$ for all $1 \leq i \leq N$ and $o \in O \setminus \{o^*\}$, 
    then $\arg \max_{o \in O} \spbwn(o) = \{o^*\}$. 
\end{description}
\end{proposition}
\begin{proof}
1. The assumptions imply that $\downi(o_1) > \downi(o_2)$ and therefore
$\pbw_{\succ_i}(o_1) > \pbw_{\succ_i}(o_2)$ for all $1 \leq i \leq N$.
Hence, $\spbwn(o_1) > \spbwn(o_2)$.
The same is true for $\spbwn$ because it is just a rescaling of $\spbw$.

2. Since all rankings are equal, the outcomes are totally ordered by $\succeq = \succeq_1$ in our setting. Hence, if $O = \{o_1, \dots, o_m\}$ and 
$o_1 \succeq o_2 \succeq \dots \succeq o_m$, then
$\pbw(o_i) = 2 \cdot \down(o_i) =  2 \cdot (m-i)$.
Hence, $\spbw(o_i) > \spbw(o_j)$ if and only if $o_i$ is ranked higher than $o_j$.
The same is true for $\spbwn$ because it is just a rescaling of $\spbw$.

3. Partial agreement implies that $\spbwn(o^*) > \spbwn(o)$
for all $o \in O \setminus \{o^*\}$, which implies the claim.
\end{proof}

\section{Experiments}
To assess the effectiveness of our approach, we conduct experiments 
on three
sets of ranking queries
from
manufacturing, finance, and medicine. 
We first describe our methodology for generating ranking queries and extracting responses 
in a semi-automatic manner (Algorithm.\ref{alg:dataset} provides an overview of the generation process). Subsequently, we will introduce the selected baseline approaches and the metrics used to assess the robustness of the aggregated answers. 
Code is available at \url{https://github.com/boschresearch/RobustLLM/}.

\subsection{Generation of Ranking Queries}

\subsubsection{Generate Symptom-Cause Matrices}

To generate ranking queries for our experiments, we first generate symptom-cause matrices, which contain information about a list of underlying critical problems and the possible symptoms we could observe.

We generate those matrices with ChatGPT. In the first step, we ask ChatGPT for a list of critical problems (causes) $\mathcal{C}$ in a specific domain with the following prompts:
\begin{tcolorbox}
    \emph{"In manufacturing, what are the critical problems that can severely impact the health and overall performance of the factory? Output a list of those problems and rank them based on degree of risk to factory."}
\end{tcolorbox}
\begin{tcolorbox}
    \emph{"What are the critical financial problems that can severely impact the health and overall performance of a company? Output a list of those problems and rank them based on degree of risk to company."}\\
    
    \emph{"What are common diseases with similar symptoms?"}
\end{tcolorbox}
The first step, gives us the possible diagnoses for the domain.
In the second step, we generate symptoms for each diagnosis with the following prompt:
\begin{tcolorbox}
\emph{"What can we observe in factory/company/human body to identify the underlying problem <the specific problem>? Output a list of indicators and rank them based on your confidence."}
\end{tcolorbox}
Similar to Section 4.1, we summarize and normalize the symptoms into a list denoted as $\mathcal{S}$, eliminating redundancy arising from synonyms and syntactic variations. Subsequently, we generate matrices as presented in 
Tables \ref{tab:manufacturing_matrix}, \ref{tab:finance_matrix}, and \ref{tab:medical_matrix} (see Appendix \ref{app:dataset}).

\subsubsection{Sample Symptom Sets}
In real-world scenarios, we have to make a diagnosis based on a 
set of symptoms. Given a list of symptoms $\mathcal{S}$ and 
diagnoses $\mathcal{D}$ for a 
particular domain, we let $s_q \subseteq \mathcal{S}$ be a subset of the symptoms, which is used in the condition description of a ranking query. For example, \{\emph{Unplanned maintenance}, \emph{Increased rework and scrap}, \emph{Increased product recalls}, \emph{Increase cost}, \emph{Increased carrying costs}\} is a subset of size 5 for the manufacturing domain. We let $s_d$ denote the set of all possible symptoms that we could observe for one specific diagnosis $d\in \mathcal{D}$.

The number of all potentially possible symptom sets (all subsets of $\mathcal{S}$) is too large for our experiments.
To find a set of reasonable size, we first quantify the uncertainty of 
symptom sets and then sample a subset of symptom sets based on their uncertainty. Intuitively, the uncertainty of a symptom set is lowest
if it uniquely identifies a diagnosis. The uncertainty is highest if all
diagnoses are compatible with the symptom set.

We use the Jaccard similarity to measure the similarity between a symptom set $s_q$ and the symptoms $s_d$ associated with diagnosis $d$:
\begin{equation}
    Sim(s_q, s_d) = \frac{|s_q\cap s_d|}{|s_q\cup s_d|}
\end{equation}
We normalize it such that, for every symptom set $s_q$, the similarity values to different diagnoses sum up to $1$:
\begin{equation}
    \overline{Sim}(s_q, s_d) = \frac{{Sim(s_q, s_d)}}{\sum_{d\in \mathcal{D}} {Sim(s_q, s_d)}}
\end{equation}

Finally, we quantify the uncertainty of symptom set $s_q$ by calculating the normalized entropy of the similarity distribution:
\begin{equation}
    U(s_q) = -\frac{1}{\log_{2}(|\mathcal{D}|)}\sum_{d\in \mathcal{D}} \overline{Sim}(s_q, s_d) \log_{2}(\overline{Sim}(s_q, s_d))
\end{equation}
Note that the entropy is always between $0$ and $\log_{2}(|\mathcal{D}|))$, hence our normalized entropy is always between $0$ and $1$.

In order to investigate our method in lower/higher uncertainty settings, we sample two types of symptom sets for each query set based on $U(s_q)$.
Since the majority of symptom sets is in the high uncertainty region,
we pick the $1000$ lowest uncertainty symptom sets for the 
low uncertainty query set. For the high uncertainty set, we focus on
sets with uncertainty between $0.7$ and $0.8$. More precisely,
the two symptom sets have been computed as follows:
\begin{itemize}
    \item low uncertainty symptom sets - $S_{low}$: we sort the potential symptom sets by normalized entropy and select the 1000 symptom sets with minimum normalized entropy. 
    \item high uncertainty symptom sets - $S_{high}$: we randomly select 1000 symptom sets with normalized entropy in the range of 0.7 to 0.8.
\end{itemize}
We visualize the uncertainty distribution of $S_{low}$ and $S_{high}$
with histograms in 
\ref{app:entropy_distribution}.

\subsection{From Symptom Sets to Ranking Queries}
We study robustness with respect to 
query and syntax
uncertainty in our experiments.
To evaluate query uncertainty, we convert symptom sets to ranking queries using the template in Figure \ref{fig:queryUncertaintyPrompt}.
\begin{figure}[t]
    \centering
    \begin{tcolorbox}
        \emph{"Given we observe <symptom 1>, <symptom 2>, ... what critical problems might exist in factory? Please output top 5 possible issues ranked by confidence without additional text."}\\
        
        \emph{"Given we observe <symptom 1>, <symptom 2>, ... what critical financial issue might we have in our company? Please output top 5 possible issues ranked by confidence without additional text."}\\
    
        \emph{"Given following symptoms: <symptom 1>, <symptom 2>, ... what disease might the patient have? Please output top 5 possible issues ranked by confidence without additional text."}
    \end{tcolorbox}
    \caption{Query templates for evaluating query uncertainty}
    \label{fig:queryUncertaintyPrompt}
\end{figure}
To evaluate syntax uncertainty,
we designed two query variants to investigate the
effect of syntactic query changes that are semantically meaningless.
In the first variant, we only replace part of the words with synonyms without changing the structure of the queries (e.g. we replace "observe" with "detect" and replace "critical problems" with "essential issues"). In the second variant, we also change the structure of the query. An an example, Figure
\ref{fig:syntaxUncertaintyPrompt} shows the variants of the 
manufacturing ranking template stated before.
\begin{figure}[t]
    \centering
    \begin{tcolorbox}
    \emph{Variant 1: "Given we detect <symptom 1>, <symptom 2>, ... what essential issues might exist in factory? Please output top 5 possible issues ranked by confidence without additional text."}\\
    
    \emph{Variant 2: "What potentially serious problems in the manufacturing may there be if we notice <symptom 1>, <symptom 2>, ... ? Please output top 5 possible issues ranked by confidence without additional text."}
\end{tcolorbox}
    \caption{Syntactic variants of the manufacturing query.}
    \label{fig:syntaxUncertaintyPrompt}
\end{figure}

\begin{algorithm}[t]
\caption{The pseudocode of ranking query generation}\label{alg:dataset}
\begin{algorithmic}
    \item /* generate symptom-cause matrices */
    \State $\mathcal{D}$ $\gets$ query ChatGPT
    \For{$d\in\mathcal{D}$}
        \State $s_q$ $\gets$ query ChatGPT
        \State add $s_q$ to a list: $\mathcal{L}_s.\text{append}(s_q)$
    \EndFor
    \State Symptom-Cause matrix $\gets$ summarize and normalize $\mathcal{L}_s$ \\
    \item /* sample symptom sets */
    \For{$s_q\in S_q$}
        \State calculate Jaccard similarity: $Sim(s_q, s_d)\gets\frac{|s_q\cap s_d|}{|s_q\cup s_d|}$
        \State normalization: $\overline{Sim}(s_q, s_d)\gets\frac{{Sim(s_q, s_d)}}{\sum_{c\in \mathcal{C}} {Sim(s_q, s_d)}}$
        \State calculate normalized entropy:
        \State $U(s_q)\gets -\frac{1}{\log_{2}(|\mathcal{C}|)}\sum_{c\in \mathcal{C}} \overline{Sim}(s_q, s_d) \log_{2}(\overline{Sim}(s_q, s_d))$
        \State add $U(s_q)$ to the list: $\text{List of indicator entropy}.\text{append}(U(s_q))$
    \EndFor
    \State Rank the $S_q$ based on the normalized entropy (from largest to smallest).
    \State $S_{low}$ $\gets$ the last 1000 indicator sets.
    \State $S_{high}$ $\gets$ randomly select 1000 symptom sets with normalized entropy in the range of 0.7 to 0.8.\\
    
    \item /* generate ranking queries from symptom sets */\\
    Convert symptom sets into ranking queries using query template.
\end{algorithmic}
\end{algorithm}




\subsection{Evaluation Protocol}
We evaluate the robustness of our approach over three batches of ranking queries i.e. manufacturing, finance and medical queries, compared with two baseline approaches. 

\subsubsection{Baselines}
\begin{itemize}
    \item \textbf{Without Aggregation}: we do not aggregate  rank answers and directly evaluate the robustness of single answers.
    \item \textbf{Average Rank}: we treat each rank preference equally and aggregate the ranks by simply averaging the ranks. Given $N$ ranks $r_1\dots r_N$ to be aggregated, the aggregation function is defined as $A(r_1\dots r_N)=\frac{1}{N}\sum_{i=1}^N r_{i}$ in this case.
\end{itemize}

\subsubsection{Evaluation Metrics}
We use \textbf{Kendall's} rank correlation coefficient ($R_{\tau}$) \cite{kendall1938new} and \textbf{Spearman's} rank correlation coefficient ($R_s$) \cite{zar2005spearman} to evaluate the robustness of the aggregated ranks. 

Let $n$ be the number of items to be ranked. Kendall's rank correlation coefficient is defined as follows:
\begin{equation}
    R_{\tau} = \frac{C-D}{\binom{n}{2}} = \frac{2(C-D)}{n(n-1)},
\end{equation}
where $C$ is the number of concordant pairs (pairs that have the same order in predicted and ground truth ranks) and $D$ is the number of discordant pairs (pairs that have different order in both ranks). A higher $R_{\tau}$ value indicates a better match between the predicted and true ranks.

Spearman's rank correlation coefficient is defined as follows:
\begin{equation}
    R_s = \frac{cov(rank1, rank2)}{\sigma_{rank1}\cdot\sigma_{rank2}},
\end{equation}
where $cov(.)$ is the covariance between two variables and $\sigma$ is the standard deviation. Similar to Kendall's tau, a higher $R_s$ value indicates a better match between the predicted and true ranks. Algorithm.\ref{alg:evaluation} illustrates our approach to evaluating ranking queries.


\begin{algorithm}[t]
\caption{The pseudocode of ranking query evaluation}\label{alg:evaluation}
\begin{algorithmic}
    \item /* Evaluate query robustness*/
    \Require $Q$
    \For{$i\gets 1\dots K$}
        \State $p_i\gets A(T(Q,N,t_i))$
        \State add $p_i$ to a list: $P$.append($p_i$)
    \EndFor
    \State $R_Q = \frac{1}{2\binom{K}{2}}\sum_{p_1, p_2\in P, p_1\neq p_2}score(p_1, p_2)$\\
    \item /* Evaluate syntax robustness*/
    \Require $Q_1\dots Q_K$
    \For{$i\gets 1\dots K$}
        \State $p_i\gets A(T(Q_i,N,t_i))$
        \State add $p_i$ to a list: $P$.append($p_i$)
    \EndFor
    \State $R_Q = \frac{1}{2\binom{K}{2}}\sum_{p_1, p_2\in P, p_1\neq p_2}score(p_1, p_2)$
    
\end{algorithmic}
\end{algorithm}

\subsection{Experiment Settings}
In our experiments, we evaluate the robustness of the answer with respect to repeated queries (query uncertainty) and syntactic changes (syntax uncertainty). Algorithm \ref{alg:evaluation} explains briefly how we evaluate the robustness of the aggregated ranks.

\subsubsection{Evaluation of Query Uncertainty}
To evaluate the robustness with respect to repeated queries, we query ChatGPT $N$ times with ranking query $Q$ at time $t_1\dots t_K$ and aggregate the answers with aggregation function $A$ to get $K$ aggregated answers. 
Note in our experiment, we specifically set $K=3$ and $N=5$ ($N=1$ when $A$ is "without aggregation", since we do not aggregate answers in this baseline). The overall robustness of the query $R_q$ is evaluated by calculating pairwise Kendall's and Spearman's rank correlation coefficient (we use $score(x,y)$ to denote the calculation of both coefficients) and averaging the coefficients. The mean values and standard deviation of all $R_q$ is reported in our results.


\subsubsection{Evaluation of Syntax Uncertainty}
We also evaluate the robustness with respect to syntactic changes, the process is very similar to evaluation of query uncertainty. The only difference is that in this case, instead of repeatedly aggregating outputs for the same $Q$, we use $K$ different ranking queries with the same semantic meaning but different syntax. In our experiment, $K=3$.





\begin{table*}[t]
    \caption{Evaluation of query uncertainty: we submit the same ranking query to ChatGPT-turbo five times and then aggregate the results. We repeat this process three times and evaluate the robustness of the three aggregated results. The temperature is set to 1, which is the default setting in the web version of ChatGPT.}
    \label{tab:query_uncertainty_results}
    \resizebox{\textwidth}{!}{%
    \begin{tabular}{ccccccccccccc}
    \hline
    \multirow{2}{*}{} &
      \multicolumn{6}{c}{\textbf{High uncertainty ranking queries}} &
      \multicolumn{6}{c}{\textbf{Low uncertainty ranking queries}} \\  \cline{1-13}
     &
      \multicolumn{2}{c}{\textbf{without aggregation}} &
      \multicolumn{2}{c}{\textbf{average rank}} &
      \multicolumn{2}{c}{\textbf{PBW (our)}} &
      \multicolumn{2}{c}{\textbf{without aggregation}} &
      \multicolumn{2}{c}{\textbf{average rank}} &
      \multicolumn{2}{c}{\textbf{PBW (our)}} \\ 
    \multicolumn{1}{c}{\textbf{Dataset}} &
      \multicolumn{1}{c}{\textbf{Kendall}} &
      \multicolumn{1}{c}{\textbf{Spearman}} &
      \multicolumn{1}{c}{\textbf{Kendall}} &
      \multicolumn{1}{c}{\textbf{Spearman}} &
      \multicolumn{1}{c}{\textbf{Kendall}} &
      \multicolumn{1}{c}{\textbf{Spearman}} &
      \multicolumn{1}{c}{\textbf{Kendall}} &
      \multicolumn{1}{c}{\textbf{Spearman}} &
      \multicolumn{1}{c}{\textbf{Kendall}} &
      \multicolumn{1}{c}{\textbf{Spearman}} &
      \multicolumn{1}{c}{\textbf{Kendall}} &
      \multicolumn{1}{c}{\textbf{Spearman}} \\ \cline{1-13}
    Manufacturing &
      0.29 (0.29) &
      0.34 (0.33) &
      0.63 (0.14) &
      0.76 (0.13) &
      \textbf{0.78 (0.09)} &
      \textbf{0.84 (0.08)} &
      0.33 (0.3) &
      0.38 (0.34) &
      0.62 (0.13) &
      0.74 (0.13) &
      \textbf{0.75 (0.09)} &
      \textbf{0.81 (0.09)} \\
    Finance &
      0.49 (0.45) &
      0.53 (0.46) &
      0.74 (0.2) &
      0.82 (0.18) &
      \textbf{0.81 (0.11)} &
      \textbf{0.86 (0.09)} &
      0.57 (0.43) &
      0.6 (0.44) &
      0.75 (0.18) &
      0.82 (0.16) &
      \textbf{0.79 (0.11)} &
      \textbf{0.84 (0.1)} \\
    Medical &
      0.54 (0.44) &
      0.59 (0.46) &
      0.75 (0.28) &
      0.81 (0.27) &
      \textbf{0.83 (0.14)} &
      \textbf{0.88 (0.13)} &
      0.56 (0.6) &
      0.58 (0.61) &
      0.67 (0.53) &
      0.7 (0.54) &
      \textbf{0.84 (0.23)} &
      \textbf{0.85 (0.23)} \\ \hline
    \end{tabular}%
    }
    
\end{table*}

\begin{table*}[t]
    \caption{Evaluation of syntax uncertainty: we submit the same ranking query to ChatGPT-turbo five times and then aggregate the results. We repeat this process for three syntactic variants and evaluate the robustness of the three aggregated results. The temperature is set to 1, which is the default setting in the web version of ChatGPT.}
    \label{tab:syntactic_uncertainty_results}
    \resizebox{\textwidth}{!}{%
    \begin{tabular}{ccccccccccccc}
    \hline
    \multirow{2}{*}{} &
      \multicolumn{6}{c}{\textbf{High uncertainty ranking queries}} &
      \multicolumn{6}{c}{\textbf{Low uncertainty ranking queries}} \\  \cline{1-13}
     &
      \multicolumn{2}{c}{\textbf{without aggregation}} &
      \multicolumn{2}{c}{\textbf{average rank}} &
      \multicolumn{2}{c}{\textbf{PBW (our)}} &
      \multicolumn{2}{c}{\textbf{without aggregation}} &
      \multicolumn{2}{c}{\textbf{average rank}} &
      \multicolumn{2}{c}{\textbf{PBW (our)}} \\ 
    \multicolumn{1}{c}{\textbf{Dataset}} &
      \multicolumn{1}{c}{\textbf{Kendall}} &
      \multicolumn{1}{c}{\textbf{Spearman}} &
      \multicolumn{1}{c}{\textbf{Kendall}} &
      \multicolumn{1}{c}{\textbf{Spearman}} &
      \multicolumn{1}{c}{\textbf{Kendall}} &
      \multicolumn{1}{c}{\textbf{Spearman}} &
      \multicolumn{1}{c}{\textbf{Kendall}} &
      \multicolumn{1}{c}{\textbf{Spearman}} &
      \multicolumn{1}{c}{\textbf{Kendall}} &
      \multicolumn{1}{c}{\textbf{Spearman}} &
      \multicolumn{1}{c}{\textbf{Kendall}} &
      \multicolumn{1}{c}{\textbf{Spearman}} \\ \cline{1-13}
    Manufacturing &
  0.25 (0.2) &
  0.29 (0.23) &
  0.27 (0.19) &
  0.32 (0.22) &
  \textbf{0.43 (0.18)} &
  \textbf{0.46 (0.19)} &
  0.31 (0.23) &
  0.35 (0.25) &
  0.31 (0.2) &
  0.37 (0.23) &
  \textbf{0.46 (0.2)} &
  \textbf{0.49 (0.21)} \\
Finance &
  0.56 (0.32) &
  0.62 (0.31) &
  0.57 (0.25) &
  0.63 (0.25) &
  \textbf{0.66 (0.17)} &
  \textbf{0.71 (0.18)} &
  0.61 (0.34) &
  0.67 (0.32) &
  0.55 (0.25) &
  0.62 (0.25) &
  \textbf{0.66 (0.16)} &
  \textbf{0.72 (0.16)} \\
Medical &
  0.64 (0.28) &
  0.71 (0.26) &
  0.71 (0.22) &
  0.78 (0.2) &
  \textbf{0.8 (0.16)} &
  \textbf{0.84 (0.15)} &
  0.83 (0.26) &
  0.85 (0.24) &
  0.84 (0.22) &
  \textbf{0.87 (0.19)} &
  \textbf{0.85 (0.18)} &
  0.86 (0.18) \\ \hline
    \end{tabular}%
    }
    
\end{table*}

\subsection{Evaluation of Query Uncertainty}
Table \ref{tab:query_uncertainty_results} presents results for query uncertainty. Our approach consistently outperforms both baselines, "without aggregation" and "average rank," across all three ranking query sets, demonstrating its superiority in both high and low uncertainty scenarios.

\subsection{Evaluation of Syntax Uncertainty}
Table \ref{tab:query_uncertainty_results} provides an overview of the outcomes pertaining to syntax uncertainty. Our approach outperform both baseline methods in the majority of scenarios examined. We observe a substantial reduction of both Kendall's and Spearman's coefficients
compared to 
Table \ref{tab:syntactic_uncertainty_results}. This suggests that syntactic variants introduce more variability.
\begin{figure}[t]
    \centering
    \includegraphics[width=0.4\textwidth]{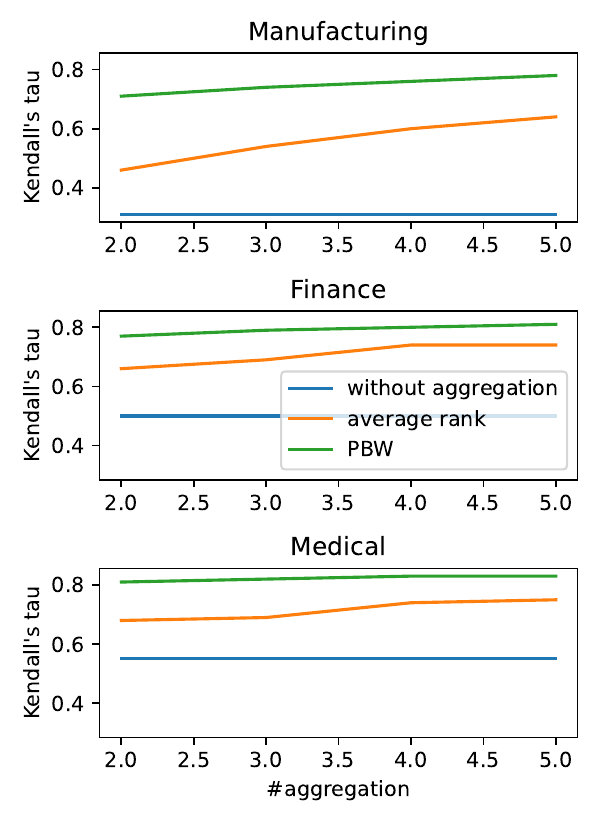}
    \caption{Robustness with respect to the number of answers used for aggregation.}
    \label{fig:num_inves}
\end{figure}

\subsection{Evaluation of Sample Efficiency
}
Another important question is how many answers do we need to aggregate? 
That is, how should we choose the parameter $N$ for our transformation method $T(Q, N, t)$. Figure \ref{fig:num_inves} shows the robustness with respect to the number of answers used for aggregation. We can see that even aggregating only two answers with our approach can already significantly increase the robustness. Note that figure \ref{fig:num_inves} shows the robustness in case of query uncertainty (high uncertainty version) and only Kendall's tau is reported. However, we observed similar trends for other settings.

\section{Conclusions}
To improve the robustness of the answers from LLMs, we suggest to 
sample answers repeatedly and to aggregate the answers using 
social choice theory. Our approach is based on the
Partial Borda Choice function 
as it gives several interesting analytical guarantees.
Our investigation primarily focuses on the application of ranking queries within diagnostic contexts, such as medical and fault diagnosis.
Our experiments show that our approach significantly improves the
robustness against both query and syntax uncertainty. 

Queries that ask for a single most plausible answer
can be understood as a degenerated special case of our ranking queries. This is because an answer $o$ can be 
understood as the partial preference $o \succ \overline{O \setminus o}$ (the provided answer is ranked
above all other answers and the ranking is indifferent about all other answers). In this special case, our average rank baseline corresponds to 
majority voting. One interesting venue for future work is to compare
partial Borda voting in the single-answer setting to other nonranking 
voting methods. One may also interpret the ranking as an
expression of approval (the answer approves of a diagnosis if it is mentioned) and to aggregate the answers by using approval voting methods.

Uncertainty in LLM outputs can also be caused by meaningless information
or adversarial attacks injected into the queries. In future work, we aim to investigate whether
social choice theory methods can also be applied to effectively
improve the robustness of LLM outputs in the presence of such perturbations.

\section{Acknowledgements}
The authors thank the International Max Planck Research School for Intelligent Systems (IMPRS-IS) for supporting Yuqicheng Zhu. The work was partially supported by the Horizon Europe projects EnrichMyData (Grant Agreement No.101070284).

\balance
\bibliographystyle{ACM-Reference-Format} 
\bibliography{sample}

\newpage
\appendix
\onecolumn
\section{Details for ranking query generation}\label{app:dataset}
\subsection{Symptom-Cause Matrices}
\begin{table}[H]
    \centering
    \begin{tabular}{|c|c|c|c|c|c|c|c|c|c|c|c|c|c|c|c|c|}
    \hline
        ~ & \rotatebox{90}{Quality Control Issues} & \rotatebox{90}{Process Inefficiencies} & \rotatebox{90}{Lack of Lean Manufacturing} & \rotatebox{90}{Machine Calibration Issues} & \rotatebox{90}{Inventory Management Issues} & \rotatebox{90}{Raw Material Shortages} & \rotatebox{90}{Supplier Quality Problems} & \rotatebox{90}{Workforce Shortages} & \rotatebox{90}{Inadequate Maintenance Programs} & \rotatebox{90}{Engineering Faults} & \rotatebox{90}{Failure to Adapt to Technological Advances} & \rotatebox{90}{Lack of Employee Training and Development} & \rotatebox{90}{Cybersecurity Threats} & \rotatebox{90}{Regulatory Compliance Violations} & \rotatebox{90}{Environmental and Sustainability Concerns} & \rotatebox{90}{Health and Safety Violations} \\ \hline
        Unexpected downtime & ~ & $\bullet$ & ~ & ~ & ~ & ~ & ~ & ~ & $\bullet$ & ~ & ~ & ~ & $\bullet$ & $\bullet$ & ~ & $\bullet$ \\ \hline
        Unplanned maintenance & ~ & $\bullet$ & ~ & ~ & ~ & ~ & ~ & ~ & $\bullet$ & ~ & ~ & ~ & $\bullet$ & $\bullet$ & ~ & ~ \\ \hline
        Reduced equipment lifespan & ~ & ~ & ~ & ~ & ~ & ~ & ~ & ~ & $\bullet$ & $\bullet$ & ~ & ~ & ~ & ~ & $\bullet$ & ~ \\ \hline
        Quality issues with products & $\bullet$ & ~ & ~ & $\bullet$ & ~ & ~ & $\bullet$ & ~ & ~ & $\bullet$ & ~ & $\bullet$ & ~ & ~ & ~ & ~ \\ \hline
        Reduced production output & $\bullet$ & $\bullet$ & ~ & $\bullet$ & ~ & $\bullet$ & ~ & $\bullet$ & ~ & $\bullet$ & $\bullet$ & $\bullet$ & $\bullet$ & ~ & ~ & ~ \\ \hline
        Delayed deliveries & ~ & $\bullet$ & ~ & ~ & ~ & $\bullet$ & $\bullet$ & $\bullet$ & ~ & $\bullet$ & ~ & ~ & $\bullet$ & $\bullet$ & ~ & $\bullet$ \\ \hline
        Increased rework and scrap & $\bullet$ & $\bullet$ & ~ & $\bullet$ & ~ & ~ & $\bullet$ & ~ & ~ & $\bullet$ & ~ & $\bullet$ & ~ & ~ & ~ & ~ \\ \hline
        increased product recalls & $\bullet$ & ~ & ~ & $\bullet$ & ~ & ~ & $\bullet$ & ~ & ~ & $\bullet$ & ~ & $\bullet$ & ~ & ~ & ~ & ~ \\ \hline
        Increased waste & $\bullet$ & $\bullet$ & $\bullet$ & $\bullet$ & ~ & ~ & $\bullet$ & ~ & ~ & ~ & ~ & $\bullet$ & ~ & ~ & $\bullet$ & ~ \\ \hline
        low quality materials & $\bullet$ & ~ & ~ & ~ & ~ & ~ & $\bullet$ & ~ & ~ & ~ & ~ & ~ & ~ & ~ & ~ & ~ \\ \hline
        less profit & ~ & ~ & ~ & ~ & ~ & ~ & ~ & ~ & ~ & ~ & ~ & ~ & ~ & ~ & ~ & ~ \\ \hline
        Increased cost & ~ & ~ & ~ & ~ & ~ & ~ & ~ & ~ & ~ & ~ & ~ & ~ & ~ & ~ & ~ & ~ \\ \hline
        Increased maintenance costs & ~ & ~ & ~ & ~ & $\bullet$ & ~ & ~ & ~ & $\bullet$ & $\bullet$ & ~ & ~ & $\bullet$ & ~ & ~ & ~ \\ \hline
        Increased production costs & $\bullet$ & $\bullet$ & $\bullet$ & $\bullet$ & ~ & $\bullet$ & $\bullet$ & ~ & $\bullet$ & $\bullet$ & $\bullet$ & ~ & $\bullet$ & ~ & ~ & ~ \\ \hline
        Increased material costs & $\bullet$ & $\bullet$ & $\bullet$ & $\bullet$ & ~ & $\bullet$ & ~ & ~ & $\bullet$ & $\bullet$ & $\bullet$ & ~ & ~ & ~ & ~ & ~ \\ \hline
        Increased energy costs & $\bullet$ & $\bullet$ & $\bullet$ & $\bullet$ & ~ & ~ & ~ & ~ & $\bullet$ & ~ & $\bullet$ & ~ & ~ & ~ & ~ & ~ \\ \hline
        Increased carrying costs & ~ & $\bullet$ & $\bullet$ & ~ & $\bullet$ & $\bullet$ & ~ & ~ & $\bullet$ & ~ & ~ & ~ & ~ & ~ & ~ & ~ \\ \hline
        increased operating cost & ~ & $\bullet$ & $\bullet$ & ~ & ~ & $\bullet$ & ~ & $\bullet$ & $\bullet$ & $\bullet$ & $\bullet$ & $\bullet$ & $\bullet$ & ~ & ~ & ~ \\ \hline
        Increased labor costs & ~ & ~ & $\bullet$ & $\bullet$ & ~ & $\bullet$ & ~ & $\bullet$ & $\bullet$ & ~ & ~ & $\bullet$ & ~ & ~ & ~ & $\bullet$ \\ \hline
        Workplace accidents and injuries & $\bullet$ & ~ & ~ & ~ & ~ & ~ & ~ & ~ & $\bullet$ & ~ & ~ & $\bullet$ & ~ & ~ & $\bullet$ & $\bullet$ \\ \hline
        Low employee morale & ~ & ~ & ~ & ~ & ~ & ~ & ~ & $\bullet$ & ~ & ~ & ~ & $\bullet$ & ~ & $\bullet$ & ~ & $\bullet$ \\ \hline
        Customer dissatisfaction & $\bullet$ & $\bullet$ & ~ & ~ & ~ & $\bullet$ & $\bullet$ & ~ & ~ & $\bullet$ & ~ & ~ & $\bullet$ & ~ & ~ & ~ \\ \hline
        reduced brand reputation & $\bullet$ & ~ & ~ & ~ & ~ & $\bullet$ & $\bullet$ & ~ & ~ & $\bullet$ & ~ & ~ & ~ & $\bullet$ & $\bullet$ & $\bullet$ \\ \hline
        Inventory shortages & ~ & $\bullet$ & $\bullet$ & ~ & $\bullet$ & $\bullet$ & ~ & ~ & ~ & ~ & ~ & ~ & ~ & ~ & ~ & ~ \\ \hline
        Obsolete inventory & ~ & $\bullet$ & $\bullet$ & ~ & $\bullet$ & ~ & ~ & ~ & ~ & ~ & ~ & ~ & ~ & ~ & ~ & ~ \\ \hline
        Inaccurate demand forecasting & ~ & ~ & $\bullet$ & ~ & $\bullet$ & ~ & ~ & ~ & ~ & ~ & ~ & ~ & ~ & ~ & ~ & ~ \\ \hline
        Frequent supplier audits & ~ & $\bullet$ & ~ & ~ & ~ & ~ & $\bullet$ & ~ & ~ & ~ & ~ & ~ & ~ & ~ & ~ & ~ \\ \hline
        Fines and penalties & $\bullet$ & ~ & ~ & ~ & ~ & ~ & $\bullet$ & ~ & ~ & ~ & ~ & ~ & ~ & $\bullet$ & $\bullet$ & $\bullet$ \\ \hline
        Increased paperwork & $\bullet$ & ~ & $\bullet$ & ~ & $\bullet$ & ~ & ~ & ~ & ~ & ~ & ~ & ~ & $\bullet$ & $\bullet$ & $\bullet$ & ~ \\ \hline
        Reduced competitiveness & ~ & $\bullet$ & ~ & ~ & ~ & ~ & ~ & ~ & ~ & ~ & $\bullet$ & ~ & ~ & ~ & $\bullet$ & ~ \\ \hline
        Limited access to certain markets & ~ & ~ & ~ & ~ & ~ & ~ & ~ & ~ & ~ & ~ & ~ & ~ & ~ & $\bullet$ & $\bullet$ & ~ \\ \hline
        Difficulty in obtaining permits & ~ & ~ & ~ & ~ & ~ & ~ & ~ & ~ & ~ & ~ & ~ & ~ & ~ & $\bullet$ & $\bullet$ & ~ \\ \hline
        Difficulty attracting skilled workers & ~ & ~ & ~ & ~ & ~ & ~ & ~ & $\bullet$ & ~ & ~ & $\bullet$ & $\bullet$ & ~ & ~ & ~ & ~ \\ \hline
        Loss of intellectual property & ~ & ~ & ~ & ~ & ~ & ~ & ~ & ~ & ~ & ~ & ~ & ~ & $\bullet$ & $\bullet$ & ~ & ~ \\ \hline
        \end{tabular}
    \caption{Symptom-Cause table for the dataset of manufacturing domain: This table outlines 16 pivotal concerns posing potential threats to manufacturing facilities (causes), along with 34 corresponding indicators (symptoms) for proactive issue identification.}
    \label{tab:manufacturing_matrix}
\end{table}

\begin{table}[H]
    \centering
    \begin{tabular}{|c|c|c|c|c|c|c|c|c|}
    \hline
         & \rotatebox{90}{Lack of Financial Planning} & \rotatebox{90}{Fraud and Embezzlement} & \rotatebox{90}{Ignoring Market Trends} & \rotatebox{90}{Overreliance on External Funding} & \rotatebox{90}{Lack of Risk Management} & \rotatebox{90}{Lack of Innovation} & \rotatebox{90}{Market Saturation} & \rotatebox{90}{Ineffective Pricing Strategies} \\ \hline
        Cash Flow Irregularities & ~ & $\bullet$ & ~ & ~ & ~ & ~ & ~ & ~ \\ \hline
        Cash Flow Variability & $\bullet$ & ~ & $\bullet$ & ~ & $\bullet$ & ~ & ~ & ~ \\ \hline
        Customer Churn & ~ & ~ & $\bullet$ & ~ & ~ & $\bullet$ & $\bullet$ & $\bullet$ \\ \hline
        Declining Sales or Revenue & ~ & ~ & $\bullet$ & ~ & $\bullet$ & $\bullet$ & $\bullet$ & ~ \\ \hline
        Decreased Market Share & ~ & ~ & $\bullet$ & ~ & ~ & $\bullet$ & $\bullet$ & $\bullet$ \\ \hline
        Decreased Marketing Expenses & ~ & ~ & ~ & ~ & ~ & $\bullet$ & $\bullet$ & ~ \\ \hline
        Delayed Payment of Suppliers & ~ & $\bullet$ & ~ & ~ & $\bullet$ & ~ & ~ & ~ \\ \hline
        Difficulty in Attracting Investors & $\bullet$ & ~ & $\bullet$ & $\bullet$ & $\bullet$ & $\bullet$ & $\bullet$ & $\bullet$ \\ \hline
        Excessive Expenses & $\bullet$ & $\bullet$ & $\bullet$ & ~ & ~ & $\bullet$ & ~ & $\bullet$ \\ \hline
        Excessive or Unauthorized Transactions & ~ & $\bullet$ & ~ & ~ & $\bullet$ & ~ & ~ & ~ \\ \hline
        Failure to Reach Financial Goals & $\bullet$ & $\bullet$ & ~ & $\bullet$ & $\bullet$ & $\bullet$ & $\bullet$ & $\bullet$ \\ \hline
        Frequent Price Adjustments & ~ & ~ & ~ & ~ & ~ & ~ & ~ & $\bullet$ \\ \hline
        High Debt Accumulation & $\bullet$ & $\bullet$ & $\bullet$ & $\bullet$ & $\bullet$ & ~ & ~ & ~ \\ \hline
        Inadequate Allocation of Resources & $\bullet$ & ~ & ~ & $\bullet$ & $\bullet$ & ~ & ~ & ~ \\ \hline
        Inadequate Provisions for Contingencies & $\bullet$ & ~ & ~ & ~ & $\bullet$ & ~ & ~ & ~ \\ \hline
        Inconsistent or Unpredictable Revenue & $\bullet$ & ~ & ~ & ~ & $\bullet$ & ~ & ~ & $\bullet$ \\ \hline
        Inefficient Cost Structure & ~ & ~ & $\bullet$ & $\bullet$ & ~ & $\bullet$ & ~ & ~ \\ \hline
        Limited Investment in Growth & $\bullet$ & ~ & $\bullet$ & $\bullet$ & ~ & $\bullet$ & $\bullet$ & ~ \\ \hline
        Limited Liquidity & ~ & $\bullet$ & ~ & $\bullet$ & ~ & ~ & ~ & ~ \\ \hline
        Lower Asset Values & ~ & ~ & ~ & $\bullet$ & $\bullet$ & $\bullet$ & ~ & $\bullet$ \\ \hline
        Misclassification of Expenses & ~ & $\bullet$ & ~ & ~ & ~ & ~ & ~ & ~ \\ \hline
        Missed Budget Targets & $\bullet$ & ~ & $\bullet$ & $\bullet$ & ~ & ~ & ~ & $\bullet$ \\ \hline
        Negative Cash Flow & $\bullet$ & ~ & ~ & $\bullet$ & $\bullet$ & ~ & ~ & $\bullet$ \\ \hline
        Obsolete Inventory & ~ & ~ & $\bullet$ & ~ & ~ & $\bullet$ & ~ & ~ \\ \hline
        Reduced Profit Margins & $\bullet$ & ~ & $\bullet$ & ~ & ~ & $\bullet$ & $\bullet$ & $\bullet$ \\ \hline
        Slower Revenue Growth & $\bullet$ & ~ & $\bullet$ & ~ & ~ & $\bullet$ & $\bullet$ & ~ \\ \hline
        Unexplained Discrepancies & $\bullet$ & $\bullet$ & ~ & ~ & $\bullet$ & ~ & ~ & ~ \\ \hline
        Unusual Accounting Entries & ~ & $\bullet$ & ~ & ~ & ~ & ~ & ~ & $\bullet$ \\ \hline
    \end{tabular}
    \caption{Symptom-Cause table for the dataset of finance domain: This table outlines 8 pivotal financial concerns posing potential threats to company (causes), along with 28 corresponding indicators (symptoms) for proactive issue identification.}
    \label{tab:finance_matrix}
\end{table}

\begin{table}[H]
    \centering
    \begin{tabular}{|c|c|c|c|c|c|c|c|c|c|c|}
    \hline
         & \rotatebox{90}{Common Cold} & \rotatebox{90}{Influenza} & \rotatebox{90}{COVID-19} & \rotatebox{90}{Allergies} & \rotatebox{90}{Acute Bronchitis} & \rotatebox{90}{Chronic Bronchities} & \rotatebox{90}{Pneumonia} & \rotatebox{90}{Sinusitis} & \rotatebox{90}{Asthma} & \rotatebox{90}{Adenovirus} \\ \hline
        Runny or Stuffy Nose & $\bullet$ & $\bullet$ & $\bullet$ & $\bullet$ & $\bullet$ & ~ & ~ & $\bullet$ & $\bullet$ & $\bullet$ \\ \hline
        Sneezing & $\bullet$ & $\bullet$ & $\bullet$ & $\bullet$ & $\bullet$ & ~ & ~ & ~ & ~ & $\bullet$ \\ \hline
        Sore Throat & $\bullet$ & $\bullet$ & $\bullet$ & $\bullet$ & $\bullet$ & ~ & ~ & $\bullet$ & ~ & $\bullet$ \\ \hline
        Cough & $\bullet$ & $\bullet$ & $\bullet$ & $\bullet$ & $\bullet$ & $\bullet$ & $\bullet$ & $\bullet$ & $\bullet$ & $\bullet$  \\ \hline
        Fever & $\bullet$ & $\bullet$ & $\bullet$ & ~ & $\bullet$ & ~ & $\bullet$ & $\bullet$ & ~ & $\bullet$ \\ \hline
        Fatigue & $\bullet$ & $\bullet$ & $\bullet$ & ~ & $\bullet$ & ~ & $\bullet$ & $\bullet$ & ~ & $\bullet$ \\ \hline
        Headache & $\bullet$ & $\bullet$ & $\bullet$ & $\bullet$ & ~ & ~ & $\bullet$ & $\bullet$ & ~ & $\bullet$ \\ \hline
        Muscle Aches & $\bullet$ & $\bullet$ & $\bullet$ & $\bullet$ & ~ & ~ & $\bullet$ & ~ & ~ & ~ \\ \hline
        Shortness of Breath & ~ & $\bullet$ & $\bullet$ & $\bullet$ & $\bullet$ & $\bullet$ & $\bullet$ & $\bullet$ & $\bullet$ & $\bullet$ \\ \hline
        Chest Pain & ~ & $\bullet$ & ~ & $\bullet$ & $\bullet$ & $\bullet$ & ~ & ~ & $\bullet$ & $\bullet$ \\ \hline
        Nausea & ~ & ~ & $\bullet$ & $\bullet$ & ~ & ~ & $\bullet$ & ~ & ~ & $\bullet$ \\ \hline
        Vomiting & ~ & ~ & $\bullet$ & ~ & ~ & ~ & $\bullet$ & ~ & ~ & $\bullet$ \\ \hline
        Diarrhea & ~ & ~ & $\bullet$ & ~ & ~ & ~ & $\bullet$ & ~ & ~ & ~ \\ \hline
        Abdominal Pain & ~ & ~ & ~ & ~ & ~ & ~ & $\bullet$ & ~ & ~ & ~ \\ \hline
        Skin Rash & ~ & ~ & $\bullet$ & $\bullet$ & ~ & ~ & ~ & ~ & ~ & $\bullet$ \\ \hline
        Chills & $\bullet$ & $\bullet$ & $\bullet$ & ~ & ~ & ~ & $\bullet$ & ~ & ~ & ~ \\ \hline
        Swollen Lymph Nodes & ~ & ~ & $\bullet$ & ~ & ~ & ~ & $\bullet$ & ~ & ~ & ~ \\ \hline
        Cognitive Impairment & ~ & ~ & $\bullet$ & ~ & ~ & ~ & ~ & ~ & ~ & ~ \\ \hline
        Weight Loss & ~ & ~ & ~ & ~ & ~ & ~ & ~ & ~ & ~ & $\bullet$ \\ \hline
        Sensitivity to Light & $\bullet$ & $\bullet$ & ~ & ~ & ~ & ~ & ~ & ~ & ~ & $\bullet$  \\ \hline
        Difficulty Concentrating & ~ & ~ & $\bullet$ & ~ & ~ & ~ & ~ & ~ & ~ & ~ \\ \hline
        Loss of Interest in Activities & ~ & ~ & ~ & ~ & ~ & ~ & ~ & ~ & ~ & ~ \\ \hline
        watery eyes & $\bullet$ & $\bullet$ & $\bullet$ & ~ & ~ & ~ & ~ & ~ & ~ & $\bullet$ \\ \hline
        Postnasal Drip & $\bullet$ & ~ & ~ & ~ & ~ & ~ & ~ & ~ & ~ & ~ \\ \hline
        Diarrhea & ~ & $\bullet$ & ~ & ~ & ~ & ~ & ~ & ~ & ~ & ~ \\ \hline
        Loss of Appetite & ~ & $\bullet$ & ~ & ~ & ~ & ~ & ~ & ~ & ~ & $\bullet$ \\ \hline
        Difficulty Sleeping & ~ & $\bullet$ & $\bullet$ & ~ & ~ & ~ & ~ & ~ & $\bullet$ & ~ \\ \hline
        Wheezing & ~ & ~ & ~ & $\bullet$ & ~ & $\bullet$ & $\bullet$ & ~ & $\bullet$ & ~ \\ \hline
        Hives & ~ & ~ & ~ & $\bullet$ & ~ & ~ & ~ & ~ & ~ & ~ \\ \hline
        Swelling & ~ & ~ & ~ & $\bullet$ & ~ & ~ & ~ & ~ & ~ & ~ \\ \hline
        Frequent Respiratory Infections & ~ & ~ & ~ & ~ & ~ & $\bullet$ & ~ & ~ & ~ & ~ \\ \hline
        Facial Pain or Pressure & ~ & ~ & ~ & ~ & ~ & ~ & ~ & $\bullet$ & ~ & ~ \\ \hline
        loss of smell & ~ & ~ & $\bullet$ & ~ & ~ & ~ & ~ & $\bullet$ & ~ & ~ \\ \hline
        loss of taste & ~ & ~ & $\bullet$ & ~ & ~ & ~ & ~ & ~ & ~ & ~ \\ \hline
    \end{tabular}
    \caption{Symptom-Cause table for the dataset of medical domain: This table outlines 10 diseases, along with 34 corresponding symptoms.}
    \label{tab:medical_matrix}
\end{table}

\twocolumn
\subsection{Entropy distribution of the ranking queries}\label{app:entropy_distribution}
\begin{figure}[H]
    \centering
    \includegraphics[width=0.45\textwidth]{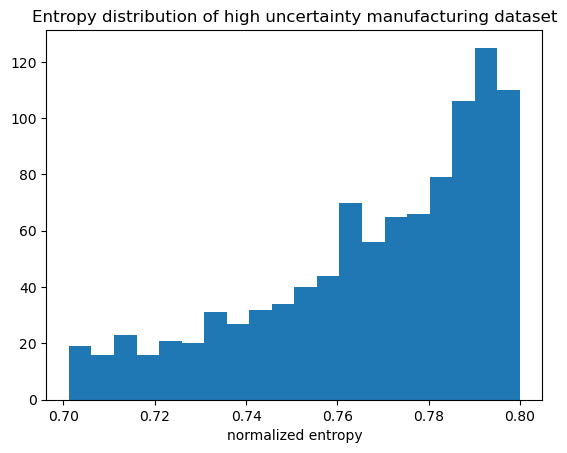}
    \label{fig:man_high}
\end{figure}
\begin{figure}[H]
    \centering
    \includegraphics[width=0.45\textwidth]{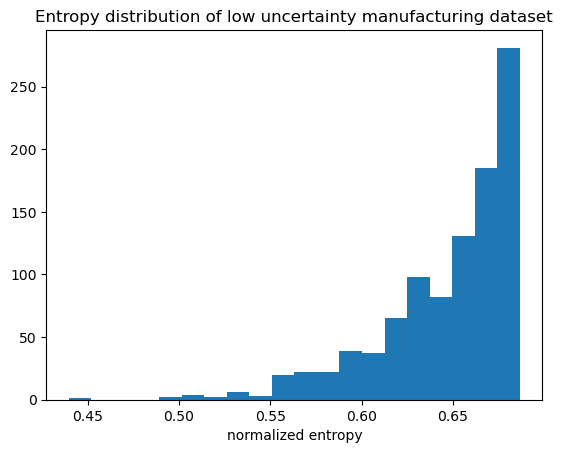}
    \label{fig:man_low}
\end{figure}
\begin{figure}[H]
    \centering
    \includegraphics[width=0.45\textwidth]{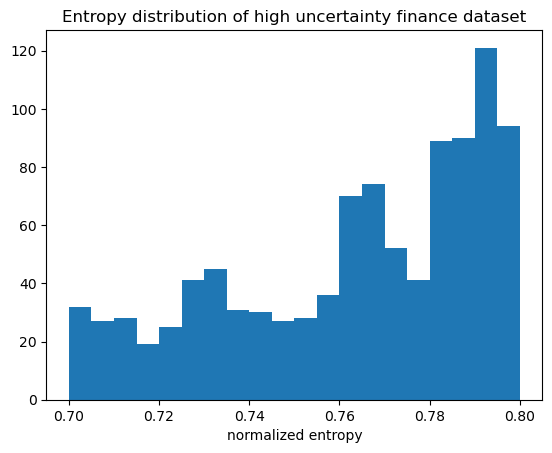}
    \label{fig:fin_high}
\end{figure}
\hfill
\begin{figure}[H]
    \centering
    \includegraphics[width=0.45\textwidth]{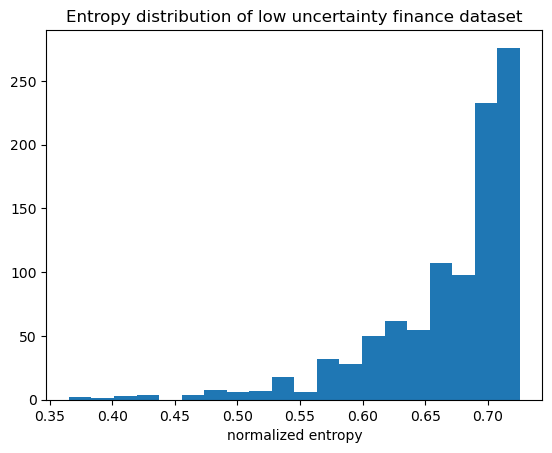}
    \label{fig:fin_low}
\end{figure}
\begin{figure}[H]
    \centering
    \includegraphics[width=0.45\textwidth]{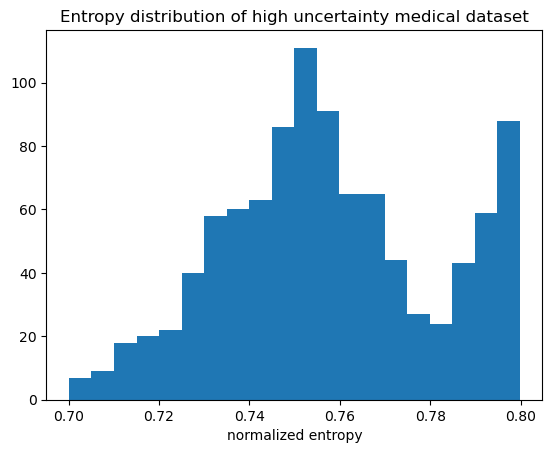}
    \label{fig:med_high}
\end{figure}
\begin{figure}[H]
    \centering
    \includegraphics[width=0.45\textwidth]{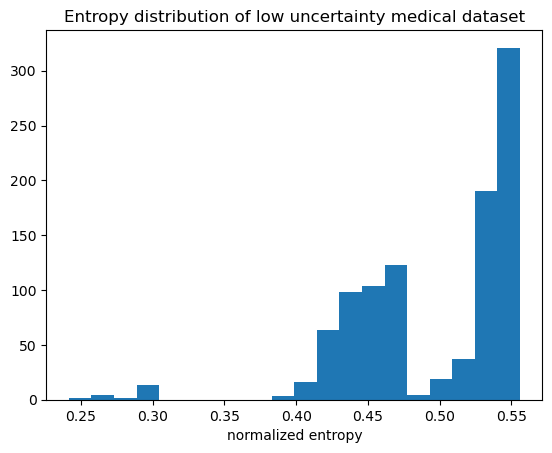}
    \label{fig:med_low}
\end{figure}


\end{document}